\pdfoutput=1

\documentclass[11pt]{article}

\usepackage[]{acl}

\usepackage{times}
\usepackage{latexsym}

\usepackage[T1]{fontenc}

\usepackage[utf8]{inputenc}

\usepackage{microtype}

\usepackage{enumitem}
\usepackage{todonotes}

\usepackage{booktabs}

\usepackage{multirow}
\usepackage{graphicx}

\usepackage{cleveref}
\crefname{section}{§}{§§}
\Crefname{section}{§}{§§}

\usepackage{gb4e} 
\noautomath

\DeclareCaptionLabelFormat{AppendixTables}{\thesection.#2}
\DeclareCaptionLabelFormat{AppendixFigure}{\thesection.#2}

\newcommand{\chiara}[1]{\textcolor{black}{#1}}

\newcommand{\sociofillmore}{\mbox{\textsc{SocioFillmore}}}
\newcommand{\fnframe}[1]{\textsc{#1}}

\title{\sociofillmore: A Tool for Discovering Perspectives}

\author{Gosse Minnema$^{1}$, Sara Gemelli$^{2}$, Chiara Zanchi$^{2}$, Tommaso Caselli$^{1}$, Malvina Nissim$^{1}$ \\
  1. University of Groningen, The Netherlands \\
  2. University of Pavia, Italy\\
  }

\begin{document}
\maketitle
\begin{abstract}
\sociofillmore\ is a multilingual tool which helps to bring to the fore the \textit{focus} or the \textit{perspective} that a text expresses in depicting an event. Our tool, whose rationale we also support through a large collection of human judgements, is theoretically grounded on frame semantics and cognitive linguistics, and implemented using the LOME frame semantic parser. We describe \sociofillmore's development and functionalities, show how non-NLP researchers can easily interact with the tool, and present some example case studies which are already incorporated in the system, together with the kind of analysis that can be visualised. 

\end{abstract}

\section{Introduction}
\label{sec:introduction}

\begin{figure}
    \centering
    \includegraphics[width=1.1\columnwidth]{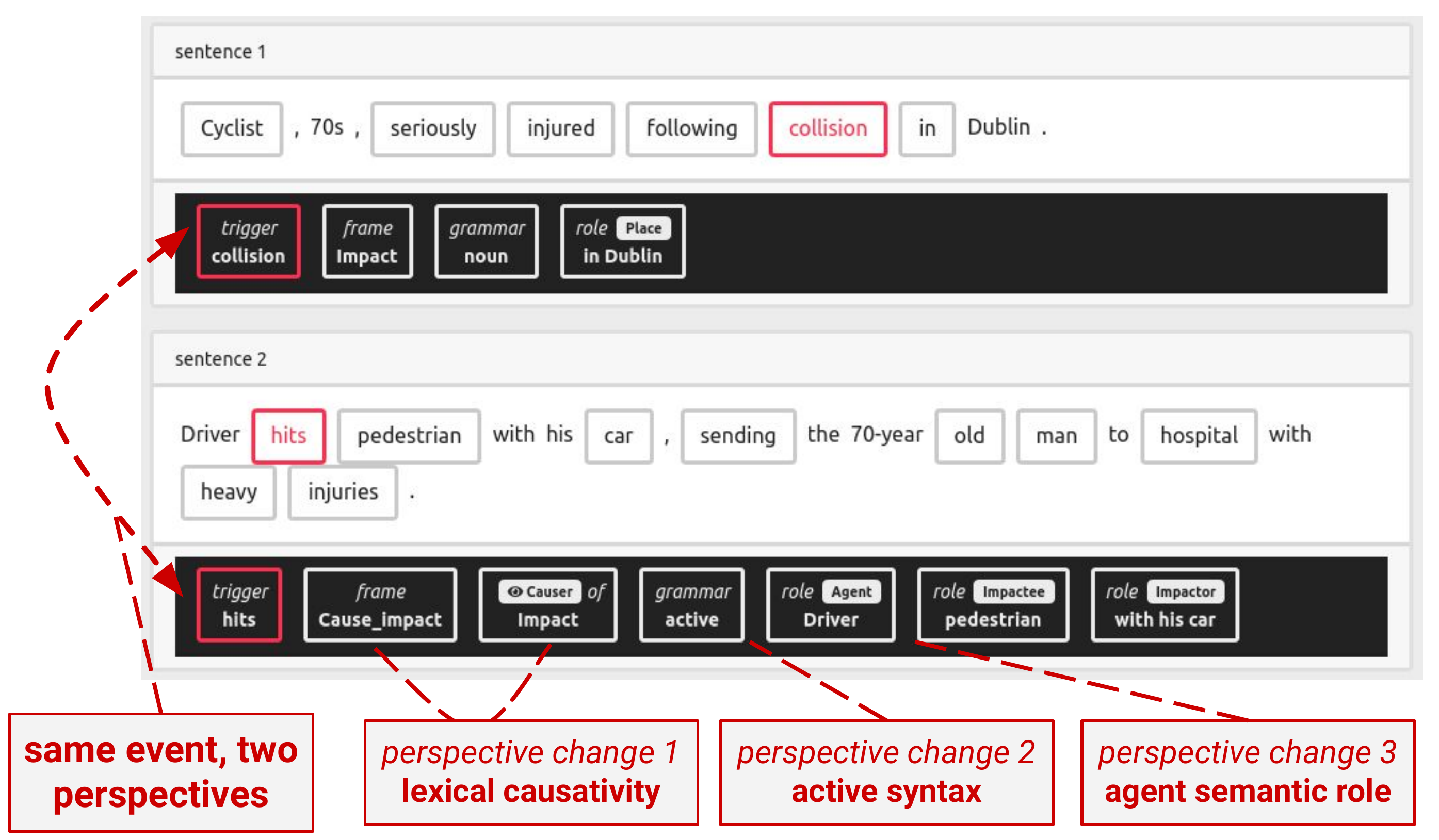}
    \caption{Analysis from \sociofillmore\ showing linguistic markers indicating the perspective changes in two descriptions of the same event. Words in boxes indicate triggers of semantic frames in the sentence.}
    \label{fig:demo-example}
\end{figure}

Descriptions of the very same event can vary widely.
Sometimes 
completely different versions of a situation can be reported, while other times 
more subtle differences emerge by the way in which such situations or episodes are depicted by choosing specific natural language expressions. Figure~\ref{fig:demo-example} illustrates this: in the first sentence the car crash is lexicalized by the noun ``\texttt{collision}'' leaving the entire dynamic unknown and suggesting that the cyclist may have some responsibility. On the contrary, the second sentence uses the verb ``\texttt{hit}''  with a subject (the agent) and an object (the patient) making more transparent how the event happened and who is responsible for it.

This phenomenon is known as \textit{framing} or \textit{perspsectivization}, and can happen in any discourse either in full awareness or, more often, unconsciously~\cite{horst2020patterns}. Politics, for instance, is the prime arena where intentional and biased framing takes place~\cite{iyengar1994anyone,entman2002framing}, but this occur also in other domains, such as sports~\cite{semetko2000framing,matthes2012framing}. 
Indeed, representation alternatives, namely the different choices that language allows to describe the same event (not only at the lexical but also at the syntactic and pragmatic level), are key to expressing and understanding the ideological power of discourse \cite{haynes1989introducing}.


Different theoretical frameworks can guide the study of framing and perspectivization in discourse, such as Critical Discourse Analysis (CDA)~\cite{fairclough2010critical,vandijk1995} and Frame Semantics \cite{fillmore1985frames,baker2003}. While NLP tools based on some of such frameworks do exist to potentially support large-scale text analysis of perspectivization, and more specifically of Fillmore's frame semantics  \cite{xia-etal-2021-lome}, they are (i) recent, thus not yet established as analysis tools for specific perspectivization problems outside of the NLP community; (ii) technical, so that their adoption is basically impossible for the non-experts, who would though benefit from them.

We fill this gap with the development of \sociofillmore, a user-friendly multilingual tool based on frame semantics which allows users to conduct large-scale analyses of text by highlighting the perspectivization strategies they adopt.\footnote{An online demo of the app will be released soon.}

\section{Context \& Evidence}

In this section, we highlight the theoretical frameworks that have guided the design of \sociofillmore and the empirical evidence we have collected 
to support its application 
to written corpora. 

\subsection{Frames, Constructions, \& Perspectives}

Being able to depict or report about events is part of the broader human ability of storytelling~\cite{boyd2009origin,gottschall2012storytelling}. One of the key properties of telling a story is the presence of a \textit{focus}, or a perspective~\cite{mieke97narratology}, which 
is embedded and intrinsic into every communication act, as shown in Figure~\ref{fig:demo-example}. 
Furthermore, 
the lexical units in a discourse are powerful access points to complex conceptual structures of encyclopedic knowledge. This vision  is at the core of Fillmore’s frame semantics~\cite{fillmore1971,fillmore1985frames,fillmore2006}, and encoded into the FrameNet project~\cite{baker2003}.

Frames are powerful devices that express perspectives 
and they strongly interact with linguistic constructions~\cite{langacker2006voice}. 
In some cases, the construction used to present the events can even lead the evocation of different frames. This is clearly illustrated in Figure~\ref{fig:demo-example}: the noun \texttt{collision} evokes the frame \textsc{impact}, while the verb \texttt{hit} in an active voice evokes the frame \textsc{cause\_impact}. The change of frames associated with different constructions triggers different perspectives and, in this case, also different responsibilities of the participants of the event.


\chiara{In cognitive linguistic terms, these different constructional options 
are named \textit{construals}~\cite{langacker1991}. Frames and construals 
are socially, culturally, and discursively constructed and constitutive~\cite{dirven2007cognitive}: they mirror our ideology, beliefs, and stereotypes and in turn contribute to building and enhancing them.} Magnifying the connections between these two elements is a way to support users (e.g., social scientist, journalists, linguists, media studies scholars, among others) to identify and study the perpetration of biases and power structures in discourse.    




\subsection{Empirical Evidence}

As a rationale for the validity of our tool two pieces of evidence are needed. The first concerns the feasibility and accuracy of (multilingual) frame semantic parsing, since any reasoning over the significance of finding one frame activated rather than an alternative one is meaningful only if frames can be accurately detected. For this, we ran an evaluation of LOME \cite{xia-etal-2021-lome}, a multilingual end-to-end frame parsing system which is the backbone of our tool, and found that it indeed produces reliable analyses. Details of the fine-tuning procedure and the evaluation are in Section~\ref{sec:components}. 

The second piece of evidence concerns the relationship between frames and perspectives from a cognitive viewpoint, and more specifically on the human perception of semantic frames and/or of the interaction between syntactic and lexical semantic expressions of agentivity. Is it true that certain frames and/or constructions are associated with agentivity more than others?
To test this, we ran a questionnaire where participants had to express judgements about their perception of the \textit{focus} (i.e., \textit{on which participant or entity is the main focus on the sentence?}) on a set of 400 sentences extracted from a large corpus on femicides in Italian (details in \S\ref{sub:femicides} and Appendix~\ref{sec:appendix}). 
Judgements are expressed on a 5 point Likert-scale for four dimensions, namely focus on: `the murderer', `the victim',  `an object' (e.g., a weapon), or `an abstract concept or emotion' (e.g., jealousy). The selected frames 
are reported in Table~\ref{tab:questionnaire}, together with the results, i.e., perception scores for each frame-construction pair (averaged over participants and sentences; every pair had approximately the same number of ratings). The highest level of focus on the murderer is found with the \fnframe{Killing} frame evoked by an active transitive construction; this makes sense, since this is the only situation in which the presence of a Killer role is required both syntactically and semantically.\footnote{Note that 
its syntactic realization appears to have a large influence on the perceived focus placed on the agent: on average, passive constructions evoking \fnframe{Killing} have a `murderer' score of almost two points lower than their active counterparts.} 
On the other hand, constructions perceived as placing a high focus on the victim are found across all frames except \fnframe{Event}.\footnote{This is consistent with the fact that, in FrameNet, the \fnframe{Event} frame does not include any `core' semantic roles apart from Place and Time, whereas all the other included frames include a Patient-like role that likely corresponds to the victim.}
While the analysis presented here is limited to a specific domain, there is a clear pattern in the perception scores of frames and construals, granting
sufficient ground for treating their automatically detected presence as a good proxy for how sentences perspectivize events in terms of 
%
foregrounding and backgrounding participants.

\begin{table}[]
\resizebox{\columnwidth}{!}{%
\begin{tabular}{@{}llrrrrr@{}}
\toprule
\multicolumn{2}{l}{\textbf{frame/construction}} & \multicolumn{1}{l}{\textbf{murderer**}} & \multicolumn{1}{l}{\textbf{victim**}} & \multicolumn{1}{l}{\textbf{object}} & \multicolumn{1}{l}{\textbf{\begin{tabular}[c]{@{}l@{}}concept /\\ emotion*\end{tabular}}} \\ \midrule
\multicolumn{2}{l}{\fnframe{Catastrophe}} &  &  &  &  \\
\textbf{} & nonverbal & \cellcolor[HTML]{FFFFFF}1.319 & \cellcolor[HTML]{A4DBC0}2.713 & \cellcolor[HTML]{FFFFFF}0.760 & \cellcolor[HTML]{FFFFFF}2.190 \\
\multicolumn{2}{l}{\fnframe{Dead\_or\_alive}} &  &  &  &  \\
\textbf{} & nonverbal & \cellcolor[HTML]{FFFFFF}1.195 & \cellcolor[HTML]{8ED1B0}3.387 & \cellcolor[HTML]{FFFFFF}1.386 & \cellcolor[HTML]{FFFFFF}1.993 \\
\textit{} & vrb:unaccusative & \cellcolor[HTML]{FFFFFF}1.983 & \cellcolor[HTML]{89D0AD}3.529 & \cellcolor[HTML]{FFFFFF}1.566 & \cellcolor[HTML]{FFFFFF}1.539 \\
\multicolumn{2}{l}{\fnframe{Death}} & \multicolumn{1}{l}{} & \multicolumn{1}{l}{} & \multicolumn{1}{l}{} & \multicolumn{1}{l}{} \\
\textit{} & nonverbal & \cellcolor[HTML]{FFFFFF}0.967 & \cellcolor[HTML]{92D3B4}3.247 & \cellcolor[HTML]{FFFFFF}1.507 & \cellcolor[HTML]{FFFFFF}1.914 \\
\textit{} & vrb:unaccusative & \cellcolor[HTML]{FFFFFF}1.867 & \cellcolor[HTML]{7CCAA4}3.921 & \cellcolor[HTML]{FFFFFF}1.690 & \cellcolor[HTML]{FFFFFF}1.286 \\
\multicolumn{2}{l}{\fnframe{Event}} &  &  &  &  \\
\textbf{} & nonverbal & \cellcolor[HTML]{FFFFFF}1.431 & \cellcolor[HTML]{FFFFFF}1.503 & \cellcolor[HTML]{FFFFFF}1.186 & \cellcolor[HTML]{FFFFFF}2.339 \\
\textit{} & vrb:impersonal & \cellcolor[HTML]{FFFFFF}1.169 & \cellcolor[HTML]{FFFFFF}2.201 & \cellcolor[HTML]{FFFFFF}1.309 & \cellcolor[HTML]{FFFFFF}1.949 \\
\multicolumn{2}{l}{\fnframe{Killing}} & \multicolumn{1}{l}{} & \multicolumn{1}{l}{} & \multicolumn{1}{l}{} & \multicolumn{1}{l}{} \\
\textit{} & nonverbal & \cellcolor[HTML]{FFFFFF}2.007 & \cellcolor[HTML]{FFFFFF}2.387 & \cellcolor[HTML]{FFFFFF}1.032 & \cellcolor[HTML]{FFFFFF}1.673 \\
\textit{} & other & \cellcolor[HTML]{FFFFFF}2.410 & \cellcolor[HTML]{FFFFFF}2.345 & \cellcolor[HTML]{FFFFFF}1.198 & \cellcolor[HTML]{FFFFFF}1.663 \\
\textit{} & vrb:active & \cellcolor[HTML]{7DCAA4}3.897 & \cellcolor[HTML]{A6DBC1}2.659 & \cellcolor[HTML]{FFFFFF}1.570 & \cellcolor[HTML]{FFFFFF}1.651 \\
\textit{} & vrb:passive & \cellcolor[HTML]{FFFFFF}1.947 & \cellcolor[HTML]{8CD1AF}3.425 & \cellcolor[HTML]{FFFFFF}1.491 & \cellcolor[HTML]{FFFFFF}1.315 \\ \bottomrule
\end{tabular}%
}
\caption{Average scores for survey question ``the main focus is on X''. Legend: `vrb' = verbal construction; `*' = differences between frame-construction pairs are significant at $\alpha=0.05$, `**' = significant at $\alpha=0.001$ (Kruskal-Wallis non-parametric H-test). Cells with a value $>2.5$ are highlighted in green.}\label{tab:questionnaire}
\end{table}

\section{SocioFillmore}
\label{sec:components}

\sociofillmore\ consists of two parts: on the back-end side, there is a series of linguistic analysis components, and on the front-end side, there is a number of components for interacting with the user.

\subsection{Linguistic Analysis Components}
\label{sec:components}

The linguistic analysis components are a combination of existing models and resources, linked by a set of rule-based bridging components.

\paragraph{LOME}

At the core of \sociofillmore\ is a frame semantic parser for annotating texts with FrameNet-based semantic frames and roles. While the rest of our architecture is agnostic as to what specific model is used, we decided to use LOME \cite{xia-etal-2021-lome} as this is (i) one of very few available models capable of producing end-to-end frame analyses (i.e., taking raw text as input, without pre-specifying predicates to annotate), and (ii) based on XLM-R, it is the only model we are aware of that supports zero-shot multilingual FrameNet analysis. Zero-shot multilingual predictions (given English-only annotated) data are very useful given the complexities of Multilingual FrameNet and the limited availability of training data in languages other than English. In effort to evaluate and improve LOME's multilingual capabilities, in \newcite{minnema2021-clicit}, we tested LOME against an existing benchmark for Italian (the 2011 Frame Labeling over Italian Texts Task [FLAIT], \citealt{evalita2011}), and experimented with several methods for exploiting the limited available training data for Italian for improving on this. Interestingly, in a zero-shot setting, LOME underperformed versus the previous state-of-the-art SVM-based model, \cite{croce-evalita11} by 24 percentage points (57\% vs 81\%) on the benchmark's frame detection task (i.e., predict semantic frames given gold predicates), but outperformed it on the frame boundary and argument classification tasks (i.e. predicting role spans and labels given predicates and frames). In our cross-lingual training experiments, we achieved best results training LOME on the concatenation of the available corpora for English and Italian, with a frame detection score much closer to the previous state of the art (77\%). However, in a small-scale manual annotation experiment on texts from the RAI femicides corpus (see \S\ref{sub:femicides}) focusing on a limited set of frames, we found that the zero-shot LOME model performed substantially better than the cross-lingually trained version, which seems to be largely due to a drop in predicate detection performance (i.e., given raw text, find all predicates that evoke frames); this can be explained by the nature of the available Italian annotations, which cover only one frame per sentence, making it hard for the model to learn which lexical units evoke frames. Thus, while being far from perfect, zero-shot LOME seems to be the best currently available option in practice for automatic frame annotation.


\paragraph{Syntactic Analysis}
The second main step of the \sociofillmore\ analysis pipeline is syntactic analysis. For each frame structure (i.e., a semantic frame together with its semantic roles) identified by LOME, we extract three types of syntactic information through a combination of a UD parse obtained with spaCy \cite{spacy}, the FrameNet database, and a set of hand-written rules: (i) syntactic construction, (ii) role-dependency links, and (iii) predicate-root status. We distinguish between several types of syntactic constructions; Table~\ref{tab:constructions} lists these constructions in increasing order of participant foregrounding (nonverbal and impersonal constructions do not require any event participant to be syntactically expressed, unaccusatives and passives require only a Patient-like argument, while actives also require an Agent-like role). 

Constructions are classified based on three criteria. First, we look at the part-of-speech tags produced by the UD parser, in order to separate non-verbal from verbal constructions. Second, if the construction is verbal, we use FrameNet for determining which core semantic roles are required for the construction: if the triggered frame is \fnframe{Event}, the construction is classified as impersonal; all other frames in FrameNet have been manually annotated as being either `active' or `non-active' based on the presence or absence of an Agent-like participant in the definition of the frame. This information is used to classify verbal constructions associated with non-active frames as unaccusative. Finally, semantically active frames are classified as instantiating either passive or active constructions based on the syntactic features taken from the dependency parse (e.g., finite or infinitive verb form, presence of passive auxiliary, etc.).\\
\indent Two additional types of extracted syntactic information are role-dependency links and predicate-root status. The former are labels that indicate how a semantic role label is expressed syntactically relative to the frame trigger, and are extracted from the dependency tree by a rule-based algorithm that traverses the dependency tree, starting from the frame trigger, until it encounters a token that is included in the role argument span, or until a pre-set number of maximum traversal steps has been reached. Some possible role-dependency links are \texttt{``Event:nsubj$\downarrow$''} (nominal subject, e.g. in \textit{the [event] \textbf{happened}}), \texttt{``Killer:*''} (self-referring link, e.g. in \textit{the [\textbf{assassin}] of JFK}), and \texttt{``Suspect:$\uparrow$--nsubj$\downarrow$''} (subject of an intermediate node, e.g. in \textit{the [prisoner] remains in \textbf{detention}}). For the purposes of perspective analysis, role-dependency links can provide a useful extra layer on top of construction information: for example, it can help us distinguish between agent-centered and event-centered nonverbal constructions (e.g. \textit{\textbf{assassin}}, with a self-referential Killer role, vs. \textit{\textbf{homicide}}, without a Killer role) or detect active constructions in which the main focus is on an inanimate cause rather than on an animate agent (\textit{the [accident] \textbf{killed} him} vs. \textit{The [murderer] \textbf{killed} him}). On the other hand, predicate-root status refers to the position of the frame trigger in the syntactic tree, and serves as a proxy for how `central' the construction is within the sentence. Verbal constructions are classified as `roots' only if they are the root nodes of the dependency tree (i.e. are the main verb in the sentence), and nonverbal constructions are classified as `roots' only if they are the the subject of the root node. The intuition behind this is that the main verbal construction in a sentence, or nonverbal constructions closely related to it, are more under focus than other constructions. For example, in \textit{the \textbf{homicide} happened} (root) versus \textit{he was arrested for \textbf{homicide} ten years later} (non-root), \textit{homicide} is foregrounded in the former but backgrounded in the latter.

\begin{table}[]
\resizebox{\columnwidth}{!}{%
\begin{tabular}{@{}lll@{}}
\toprule
\textbf{Construction} & \textbf{\begin{tabular}[c]{@{}l@{}}Semantic Roles\end{tabular}} & \textbf{Examples} \\ \midrule
\multirow{2}{*}{non-verbal} & \multirow{2}{*}{none} & \textit{The *murder* of MLK} \\
 &  & \textit{A *deadly* accident} \\
 &&\\
\multirow{2}{*}{vrb:impersonal} & \multirow{2}{*}{\begin{tabular}[c]{@{}l@{}}none \\  {[}Event-like{]}\end{tabular}} & \textit{It *rained*} \\
 &  & \textit{The event *occurred*} \\
 &&\\
\multirow{2}{*}{vrb:unaccusative} & \multirow{2}{*}{Patient-like} & \textit{The victim *died*} \\
 &  & \textit{He *fell* off the stairs} \\
 &&\\
\multirow{2}{*}{vrb:passive} & \multirow{2}{*}{Patient-like} & \textit{She *was found* in her house} \\
 &  & \textit{The cyclist *was hit* by a car} \\
 &&\\
\multirow{2}{*}{vrb:active} & \multirow{2}{*}{\begin{tabular}[c]{@{}l@{}}Agent-like \\ {[}Patient-like{]} \end{tabular}}  & \textit{The girl *walked* to school} \\
 &  & \textit{The police *arrested* the man} \\ \bottomrule
\end{tabular}%
}
\caption{Syntactic construction types used by \sociofillmore. Semantic roles between square braces are mandatory in a subset of constructions within the type. `vrb'=verb-based constructions}
\label{tab:constructions}
\end{table}


\subsection{User Interaction}

\begin{figure}
    \centering
    \includegraphics[width=\columnwidth]{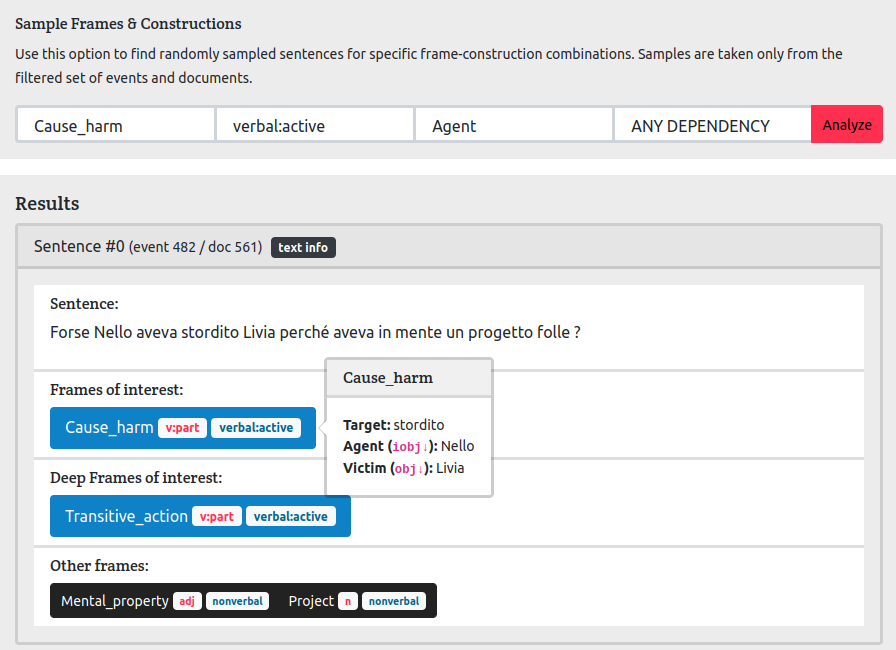}
    \caption{Explorer mode: visualize sentences matching specific linguistic features}
    \label{fig:frame-samples}
\end{figure}

\sociofillmore\ has two usage modes: one for exploring and analyzing existing, pre-processed corpora, and one for interactively exploring frame-based perspective analysis. The former mode is targeted towards domain experts interested in in-depth analysis of specific phenomena, while the latter is targeted to a broader community of (social) scientists who are unfamiliar with frame semantics but would like to learn about using linguistic frames and constructions for analyzing how events can be framed through language.

\paragraph{Corpus Explorer}

The corpus explorer is illustrated in Figure~\ref{fig:frame-samples}. It consists of three parts: first, the user can select relevant semantic frames from a pre-defined set (or add additional frames) that correspond to the perspective-taking phenomena that they want to investigate. 
Second, the user can specify which subset of the corpus they would like to analyze by adding document-based and event-based filters. Document-based filters (e.g. publication date, news outlet) can be added for any kind of corpus, while event-based filters (e.g. location of event, participants involved) can only be used for corpora that provide event-level metadata (i.e., each document is linked to some structured event representation). This second type of corpus is especially attractive for perspective analysis because it allows for investigating how similar real-world events are conceptualized in different documents that reference these events. Finally, having selected frames and filtered the corpus, the user can analyze the corpus in two ways: one can get global descriptive statistics over the selected part of the corpus, or visualize annotated sentences from the corpus. A wide range of descriptive statistics are available, for example, simple frequencies of semantic frames and constructions, frequencies of role dependencies per frame, and frame frequencies plotted as a function of time elapsed between the event occurrence and publication of the documents referencing it. On the other hand, when visualizing annotations from the corpus, there are two options: the user can either select specific documents from the corpus and analyze them sentence-by-sentence, or make a selection of linguistic features of interest (e.g. a combination of frames, constructions, and role dependencies) and request to see randomly sampled sentences matching these features.\\
\indent As of now, we have implemented the corpus explorer for four different corpora on three domains in two languages (femicides and migration in Italian, and traffic crashes in Dutch). We are planning to add additional corpora in the future, and also welcome contributed corpora from others. Adding an additional corpus requires some amount of expertise in NLP and FrameNet as well as in the domain of interest, and involves pre-processing the corpus and its metadata, running LOME and the \sociofillmore\ linguistic pipeline over the corpus, and adding corpus-specific logic (e.g. document/event filters) to the explorer interface. For future work, we are planning to develop a graphical UI to streamline this process and make it more accessible for users without a technical background.

\begin{figure}
    \centering
    \includegraphics[width=\columnwidth]{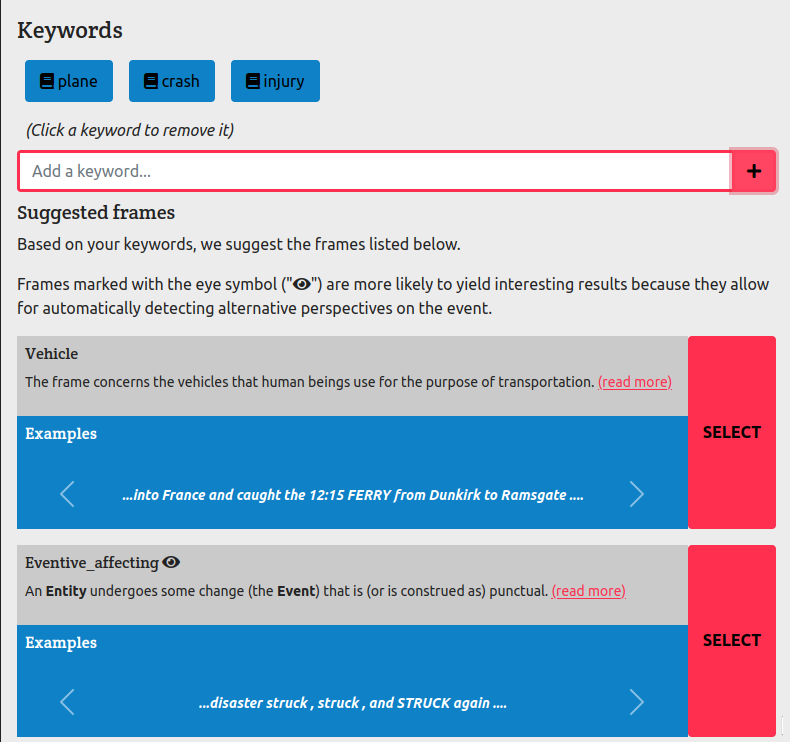}
    \caption{Interactive mode: keyword search interface}
    \label{fig:keywords}
\end{figure}

\paragraph{Interactive Mode}
The interactive mode of \sociofillmore\ aims to make frame-based perspective analysis more accessible to people without a specific background in frame semantics. To this end, we provide three main features, as shown in Figure~\ref{fig:keywords}: a step-by-step interface with examples guiding the user through the stages of perspective analysis (event definition, frame selection, and document visualization), an interactive frame selection tool, and an on-demand version of LOME and the \sociofillmore\ linguistic pipeline with simplified annotation visualization. The most novel of these features is the interactive frame selection tool, which is meant to help users who are not familiar with the FrameNet database. 
The frame selection tool consists of two components: an embedding-based keyword search function, and a rule-based algorithm for automatically finding frames that provide alternative perspectives on selected frames. The former of these makes use of `bag-of-LU' frame embeddings that are computed using a similar method to that proposed in \newcite{alhoshan-etal-2019-semantic}.
Keyword searches are performed by retrieving GloVe vectors for the specified keywords, and then finding the frame embeddings with the top-N closest cosine distance to the centroid of the set of keyword embeddings. The suggested frames are then displayed to the user along with their definition and example sentences retrieved using the NLTK FrameNet API \cite{bird2009natural}. Complementary to the keyword search system, we use frame-to-frame relations to automatically add additional frames that provide alternative perspectives on the frames that the user specified.

\section{Case studies}
\label{sec:case-studies}

\sociofillmore\ has been applied (so far) to three case studies: Italian news reporting on femicides, Italian news reporting on migration, and Dutch news reporting on traffic crashes. In each of these cases, we target events where there is a potential imbalance of power between the actors involved and the attribution of responsibility for the happening to (at least) one of the participants of the event. This makes the study of the perspectives associated with the reporting of these events very suitable to investigate how responsibility is framed in news reports of such events, and where potential representation biases may emerge.


\subsection{Femicides}
\label{sub:femicides}

For femicides, the domain of \sociofillmore\ developed most extensively to date, two corpora are currently available in the exploration tool. The first has been compiled by the CRITS research team at RAI (Radiotelevisione Italiana), composed by 2,734 news articles from 31 different Italian news sources, reporting on 937 femicides perpetrated between 2015 and 2017~\cite{belluati2021femminicidio}. The corpus is enriched with metadata (time, news source, etc.) and for each femicide event multiple news articles are available. For our analysis we selected 15 frames based on the examples in~\citet{pinelli2021gender} - see Appendix~\ref{sec:appendixb}. 

\begin{figure}
    \centering
    \includegraphics[width=\columnwidth]{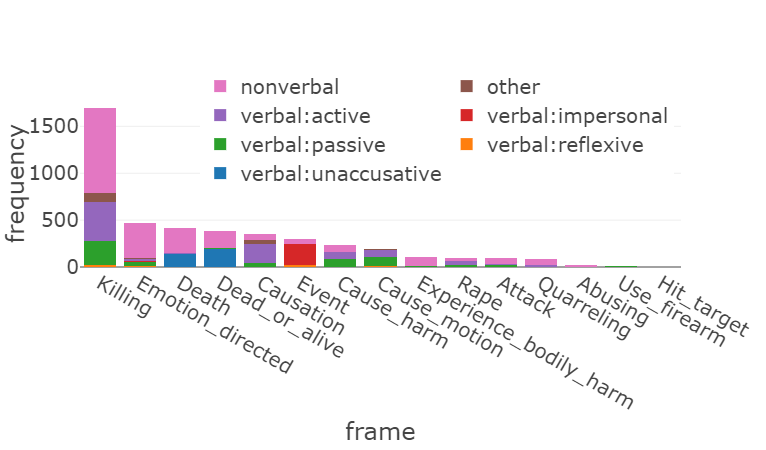}
    \caption{Femicides analysis: frequency of frames split by syntactic construction}
    \label{fig:femicide-frames-cxs}
\end{figure}

In \newcite{minnema2021-clicit}, we applied \sociofillmore\ on a randomly chosen 200K word subcorpus (10\% of all events) of the RAI corpus. The main findings are shown in Figure~\ref{fig:femicide-frames-cxs}. As expected, \textsc{Killing} is by far the most frequent typical frame, followed by \textsc{Emotion\_directed} and \textsc{Death}. Concerning syntax, the \textit{nonverbal} constructions (i.e., the predicate is either a noun or an adjective) are dominant across many frames, while \textit{verbal:active} constructions are much rarer, as well as \textit{verbal:passive} and \textit{verbal:unaccusative}. 

The combination of syntactic constructions and semantic frames is the key to magnifying perspectives. In particular, we observe that 60\% of the instances of the frame \textsc{Killing} are associated with constructions that foreground the victims and background the perpetrators. This reaches 79\% of cases when the frame used to present the event is \textsc{Death}. In terms of perspective analysis, this indicates a bias in framing femicides as events where a killing takes place but no one is actually responsible for it.

\subsection{Other domains}
\sociofillmore\ has been productively used as well for studying the framing of traffic crashes and migrations in the Dutch and Italian media, respectively. \newcite{cartadiroma2021} used the tool for automatically identifying frames contributing to either dehumanizing or humanizing migrants in newspaper headlines (e.g. reporting on `waves' of migrants, and thus collectively conceptualizing them as a non-human mass entity, vs. reporting on migrants as single and intentional individuals), and also compared the change in these frames over time relative to statistics about newly arrived migrants in Italy. We are also involved in ongoing work, in collaboration with the author of the original paper, aiming at reproducing the findings on traffic framing reported in \newcite{te2020framing}.

\section{Conclusions}

\sociofillmore\ is a multilingual tool that we have developed for studying perspectives in written text, grounded in Frame Semantics and Cognitive Linguistics. Through an interactive mode, the tool is easily accessible to non-experts, too. We support the rationale for the validity of our tool through a rigorous evaluation of the frame semantic parser at the core of our tool and a collection of human judgement on the connection between frames and perspectivization. The tool is available as a web interface (and as a docker release), it supports multiple languages, and already integrates a few large-scale case studies which can be browsed 
for research, 
but also for further understanding of the tool's functionalities.


\newpage
\section*{Ethical Statement}

One of the key properties of \sociofillmore\ is its being agnostic on whether a perspective should be considered ``good'' or ``bad''. In this respect \sociofillmore\ is \textbf{not a prescriptive tool} on how news should be reported but rather a support tool that helps to magnify misuse of frames and biases that may mirror and strengthen asymmetric power dynamics existing in our societies. 

\sociofillmore\ is based on state-of-the-art NLP technologies. While these tools achieve very good performances (also in zero-shot multilingual settings), none of them can be considered to reach nor mimic humans. The tools are based on powerful machine learning algorithms but they are far away from being artificial  intelligent agents. We recommend \textbf{caution} when using SocioFillmore since margins of errors are present. At the same time, additional tests are needed before deploying \sociofillmore\ as an integrated tool or service that citizens or professionals may use for purposes other than research.

One of the services of \sociofillmore\ is corpus-assisted language analysis. The outcome of this service is highly sensitive to the data that are feeded to the tool. This requires users to pay particular attention to the curation of the data that will compose their corpus. Results on the presence of frames and bias are a direct consequence of what is input to the tool. The case studies we have illustrated are based on carefully curated corpus collections conducted by experts. We thus recommend that users of \sociofillmore\ should accompany the presentation of their results with a \textbf{documentation of their data collections} using tools such as data statements~\cite{bender2018data} or data sheets~\cite{gebru2021datasheets}.

\bibliography{anthology,custom}
\bibliographystyle{acl_natbib}

\appendix

\renewcommand\thefigure{\thesection.\arabic{figure}}
\setcounter{figure}{0}    

\renewcommand\thetable{\thesection.\arabic{table}}
\setcounter{table}{0}

\section{Questionnaire}
\label{sec:appendix}

 The questionnaire has been conducted using the platform Qualtrics. 

 Participants have been recruited from several universities in Italy. Each participant was presented with 50 sentences and was asked to express a judgements on Likert-type scale from 0 to 5 for the perceived ``amount'' of focus placed on four dimensions, namely: `the murderer', `the victim',  `an object' (e.g., a weapon), or `an abstract concept or emotion' (e.g., jealousy). 

The sentences were selected by first automatically annotating the corpus with semantic frames and construals, determining a set of semantic frames that correspond to different ways of conceptualizing the femicides, selecting the set of most frequent construals for each frame, and then, for each frame-construal pair, randomly sampling sentences containing at least one instance of the pair from the corpus.

The frames that we selected were, in order of increasing level of detail and presence of event participants:
\begin{itemize}
\item \fnframe{Event}, e.g., \textit{the \textbf{incident} occurred};
\item \fnframe{Catastrophe}, e.g., \textit{the \textbf{tragedy} cost the life of ...}; 
\item \fnframe{Dead\_or\_alive}, e.g., \textit{the victim was found \textbf{dead}};
\item \fnframe{Death}, e.g., \textit{the victim \textbf{died} at the hands of...}; and 
\item \fnframe{Killing}, e.g., \textit{the man \textbf{murdered} his wife}. 
\end{itemize}

Each of these frames can occur in various syntactic configurations, but only \fnframe{Killing} can be evoked by a transitive verbal construction (actively or transitively used). 

While it is possible that, in some cases, instances of these frames refer to another event referenced in the texts (e.g. \fnframe{Killing} could also refer to the perpetrator committing suicide, or to some other type of secondary murder; \fnframe{Event} could also refer to other type of `events' or `incidents' mentioned in the text), but, from a manual inspection of the data, this seems to be fairly rare. We informed participants of the possibility of anomalous sentences occurring in the survey and instructed them to mark these as `irrelevant' and to not assign any points to them. 

The questionnaire has been approved by the Ethical Board of the University of Groningen. Participants were compensated with 5 euro. Each participant was asked to provide judgments on a set of 50 sentences. Participation is anonymized (there is no link between the questionnaire answers and the participants.) and we limit the collection of personal data to last name, initials, bank account and address only for payment purposes. After the payment, all personal data of the participant is deleted. 

\begin{figure}[!h]
    \centering
    \includegraphics[width=\columnwidth]{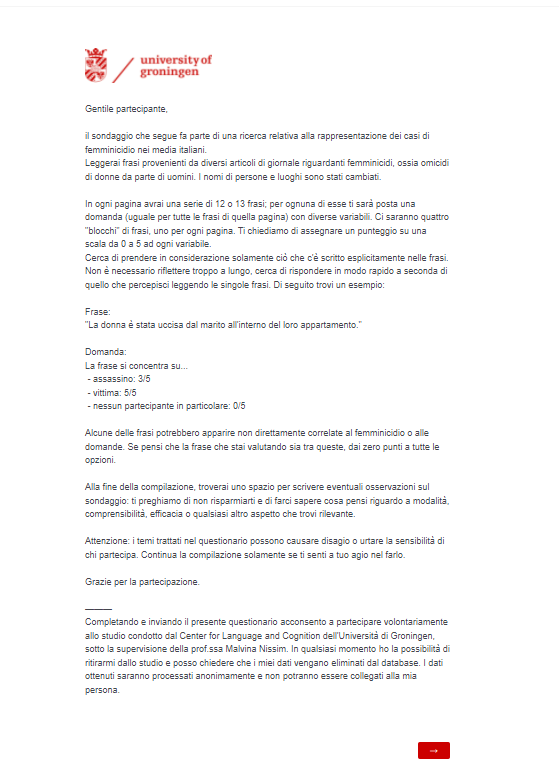}
    \caption{Original instruction given to the participant in the questionnaire for validating \sociofillmore.}
    \label{fig:demo-instructions}
\end{figure}

A screenshot of the original instructions provided to the participants is presented in Figure~\ref{fig:demo-instructions}. Translations in English of the instructions is given below:

\texttt{\textsc{instructions:}}

\texttt{Dear participant,}

\texttt{the following questionnaire is part of a project related to the representation of femicides in the italian media.
You will read sentences extracted from newspaper articles about femicides, i.e. murders in which a man kills a woman. The real names un the sentences have been changed.}

\texttt{In each page you will be presented with a series of 12 or 13 sentences; there are four different "stacks" of sentences, one for each page. For each sentence you will be asked a question (the same of every sentence in that page) followed by different variables. You will be asked to rate every variable on a scale from 0 to 5.}

\texttt{While doing the task, try to consider just what the sentence explicitly expresses. There is no need to think for too long, try to answer fast according to what you perceive reading the sentences.
Example: \textit{The woman was killed by her husband at home}. Question: The sentence focuses on...}

\begin{itemize}
\item\texttt{the murder: 3/5}
\item\texttt{the victim: 5/5}
\item\texttt{nobody in particular: 0/5}
\end{itemize}

\texttt{After the data analysis, the participants will receive an e-mail with a link to an online meeting in which we will briefly present the outcomes of the research.}

\texttt{Warning: the topics in the questionnaire could make you feel uncomfortable; please continue only if you feel at ease.
Thank you for your time!}

\texttt{By filling in the following questionnaire I consent to participate voluntarily in the study conducted by the Center for Language and Cognition of the University of Groningen, supervised by prof. dr. Malvina Nissim. I can withdraw my participation at any time and have the data obtained through this study returned to me, removed from the database or deleted. The data obtained during this study will be processed anonymously and will therefore not be able to be traced back to me.}

\section{Frames and semantic roles}
\label{sec:appendixb}
\begin{table*}[h!]
\begin{tabular}{@{}llll@{}}
\toprule
\textbf{frame} & \multicolumn{1}{c}{\textbf{role:perpetrator\_like}} & \multicolumn{1}{c}{\textbf{role:victim\_like}} & \textbf{role:cause\_like} \\ \midrule
Abusing & Abuser & Victim & - \\
Attack & Assailant & Victim & - \\
Causation & Causer & Affected & Cause \\
Cause\_harm & Agent & Victim & Cause \\
Cause\_motion & - & - & - \\
Dead\_or\_alive & - & Protagonist & Explanation \\
Death & - & Protagonist & Cause \\
Emotion\_directed & - & - & - \\
Event & - & - & - \\
Experience\_bodily\_harm & Experiencer|Body\_part & - & - \\
Hit\_target & Agent & Target & - \\
Killing & Killer & Victim & Cause \\
Quarreling & - & - & - \\
Rape & Perpetrator & Victim & - \\
Use\_firearm & Agent & Goal & - \\ \bottomrule
\end{tabular}
\caption{Femicides: mapping frames, participants, and roles}\label{tab:femicides-mappings}
\end{table*}


\newpage

In Tables~\ref{tab:femicides-mappings}, 
we provide the set of semantic frames and a mapping between their associated semantic roles and the main participants in the femicides.




\end{document}